\documentclass[10pt,twocolumn,letterpaper]{article}

\usepackage{cvpr}              
\usepackage{tabularx}
\usepackage{multirow} 
\usepackage{array}       
\usepackage{booktabs}
\usepackage[table]{xcolor}
\usepackage[pagebackref,breaklinks,colorlinks, allcolors=cvprblue]{hyperref}
\definecolor{cvprblue}{rgb}{0.21,0.49,0.74}
\def\paperID{3547} 
\def\confName{CVPR}
\def\confYear{2026}

\title{	HVG-3D: Bridging Real and Simulation Domains for 3D-Conditional Hand-Object Interaction Video Synthesis}

\author{
Mingjin Chen\textsuperscript{1 *} \quad
Junhao Chen\textsuperscript{3 *} \quad
Zhaoxin Fan\textsuperscript{2 \textdagger} \quad
Yujian Lee\textsuperscript{4} \quad
Zichen Dang\textsuperscript{1}\\
Lili Wang\textsuperscript{5} \quad
Yawen Cui\textsuperscript{1} \quad
Lap-Pui Chau\textsuperscript{1} \quad
Yi Wang\textsuperscript{1 \textdagger}\\
\\
\textsuperscript{1}Dept. of EEE, The Hong Kong Polytechnic University \quad
\textsuperscript{2}Beijing Advanced Innovation Center\\for Future Blockchain and Privacy Computing, School of Artificial Intelligence, Beihang University \\ 
\textsuperscript{3}Tsinghua University \quad
\textsuperscript{4}Beijing Normal-Hong Kong Baptist University \quad
\textsuperscript{5}State Key Laboratory of\\Virtual Reality Technology and Systems, School of Computer Science and Engineering, Beihang University
}
\begin{document}
\setcounter{page}{1}
\pagenumbering{arabic}
\pagestyle{plain}
\twocolumn[{
\renewcommand\twocolumn[1][]{#1}
\maketitle
\begin{center}
    \vspace{-7mm}
    Project Page: \url{https://hvg3d.github.io} \\
    \vspace{-2mm}
\end{center}

\begin{center}
    \captionsetup{type=figure}
    \centerline{\includegraphics[width=\linewidth]{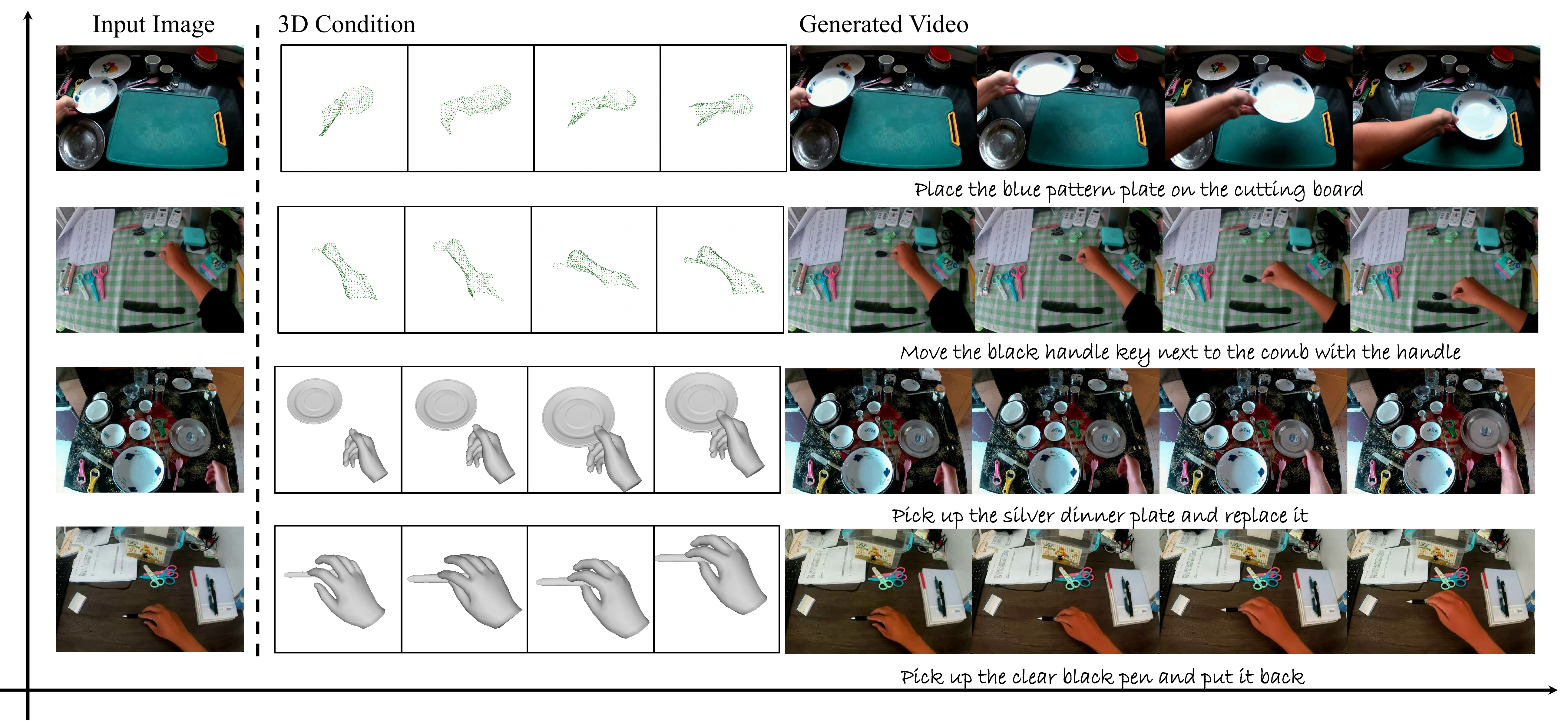}}
    \caption{
    Illustration of 3D-conditioned hand–object interaction video generation with our proposed \textbf{HVG-3D} framework. HVG-3D synthesizes realistic and temporally coherent hand–object interaction videos by conditioning on explicit 3D signals. The top two rows display generated results using 3D point cloud and pose conditions extracted from real-world egocentric videos. The bottom two rows show results where 3D conditions are obtained from simulated hand–object sequences, demonstrating the framework’s flexibility in accepting both real and synthetic 3D inputs. For each example, the leftmost column shows the input image and 3D condition, while subsequent columns depict selected frames from the generated video.
    }
    \label{fig:teaser}

\end{center}
}]

\renewcommand{\thefootnote}{} 
\footnotetext{
    \textsuperscript{*} Equal Contribution \quad
    \textsuperscript{\textdagger} Corresponding Author.
}

\setcounter{footnote}{0}
\renewcommand{\thefootnote}{\arabic{footnote}} 

\begin{abstract}
Recent methods have made notable progress in the visual quality of hand-object interaction video synthesis. However, most approaches rely on 2D control signals that lack spatial expressiveness and limit the utilization of synthetic 3D conditional data. To address these limitations, we propose HVG-3D, a unified framework for 3D-aware hand-object interaction (HOI) video synthesis conditioned on explicit 3D representations. Specifically, we develop a diffusion-based architecture augmented with a 3D ControlNet, which encodes geometric and motion cues from 3D inputs to enable explicit 3D reasoning during video synthesis. To achieve high-quality synthesis, HVG-3D is designed with two core components: (i) a 3D-aware HOI video generation diffusion architecture that encodes geometric and motion cues from 3D inputs for explicit 3D reasoning; and (ii) a hybrid pipeline for constructing input and condition signals, enabling flexible and precise control during both training and inference. During inference, given a single real image and a 3D control signal from either simulation or real data, HVG-3D generates high-fidelity, temporally consistent videos with precise spatial and temporal control. Experiments on the TASTE-Rob dataset demonstrate that HVG-3D achieves state-of-the-art spatial fidelity, temporal coherence, and controllability, while enabling effective utilization of both real and simulated data.

\end{abstract}    
\section{Introduction}
%

Recent breakthroughs in diffusion-based generative models have fundamentally advanced the field of video synthesis, with large-scale models such as Sora~\cite{brooks2024video}, CogVideo-X~\cite{yang2024cogvideox}, Keling~\cite{kuaishou2024kling}, Hunyuan Video~\cite{kong2024hunyuanvideo}, and Veo 3~\cite{google2024veo3} setting new standards for generating high-quality and temporally consistent videos. Leveraging the capabilities of these foundational models, a growing body of work has focused on the generation of hand-object interaction videos, which has garnered increasing interest for applications in training robotic grasping models~\cite{maitin2010cloth, domae2014fast, zhong2025dexgrasp, wang2025dexh2r, zhu2025evolvinggrasp, mandikal2022dexvip, brohan2022rt, zitkovich2023rt, black2410pi0, intelligence2025pi05visionlanguageactionmodelopenworld, duan2025robopara}.


However, while recent methods for hand-object interaction video generation~\cite{akkerman2025interdyn, sudhakar2024controllingworldsleighthand, dang2025svimo, zhao2025taste, pang2025manivideo, zhang2024hoidiffusion} have demonstrated impressive visual quality, their reliance on 2D conditioning signals remains a fundamental bottleneck. In particular, widely adopted controls, such as point trajectories~\cite{zhou2025trackgo, wang2024motionctrl}, optical flow~\cite{li2025image, shi2024motion, yin2023dragnuwa, zhang2025tora, lee2025optical}, bounding boxes~\cite{jain2024peekaboo, namekata2024sg, qiu2024freetraj, wang2024boximator}, and masks~\cite{akkerman2025interdyn, wang2024vividpose, tu2025stableanimator, chen2025dancetogether}, are inherently limited in spatial expressiveness and temporal consistency. This absence of true 3D conditioning introduces two critical challenges: (1) Imperfect 3D Understanding: 2D signals provide only partial motion and geometry cues, frequently resulting in unrealistic deformations and physically implausible hand-object interactions; (2) High Data Cost: These 2D conditions are typically extracted from real-world videos, making it difficult to exploit synthetic data generated by efficient simulators, and thus substantially increasing the cost of data collection and annotation.

To address the issue, recent work such as Diffusion as Shader (DaS)~\cite{gu2025diffusion} has begun to incorporate 3D tracking videos for richer motion guidance. Nevertheless, these 3D cues are ultimately projected into 2D video sequences for model input, which prevents full utilization of the spatial structure and depth relations intrinsic to 3D space. To overcome these limitations, it is essential to design methods that can intrinsically exploit 3D conditioning, thereby improving the realism and physical plausibility of hand-object interactions, as well as facilitating scalable data generation using simulators.

To this end, we present HVG-3D, a unified framework that enables 3D-aware synthesis of hand-object interaction videos. Our key insight is to bridge the gap between visual realism and precise physical control by conditioning video generation on explicit 3D representations. Given a single real-world RGB image as appearance input and a 3D condition derived from a simulator (or collected from another real videos), HVG-3D is capable of generating high-fidelity, temporally consistent hand-object interaction videos.    Specifically, the HVG-3D framework is composed of two key components. First, a 3D-aware HOI video generation diffusion architecture leverages a dedicated 3D ControlNet to encode geometric and motion cues from 3D point clouds or tracking sequences, injecting these features into a diffusion transformer via zero-initialized convolutional layers for explicit 3D reasoning. Second, a hybrid pipeline constructs input and condition signals by pairing real images with 3D conditions from simulation or another videos, supporting flexible and precise control throughout both training and inference. Both design together enable HVG-3D to generate realistic, 3D-consistent hand-object interaction videos from a single real image and a 3D control signal, as inllsurated in Fig. \ref{fig:teaser}.

Extensive experiments on the TASTE-Rob dataset demonstrate that HVG-3D significantly outperforms state-of-the-art methods across multiple metrics, achieving superior spatial fidelity, temporal coherence, and controllability, highlighting its practical value for scalable and controllable video generation. Our contributions can be summarized as:

\begin{itemize}
\item We introduce a practical paradigm for hand-object interaction video generation that bridges real and simulated domains, enabling synthesis from a real input image and a 3D condition obtained from either simulation or another real video.
\item We present HVG-3D, a unified framework featuring a 3D-aware diffusion-based architecture and a hybrid pipeline for constructing input and condition signals, achieving flexible and precise control.
\item We validate our approach with comprehensive experiments, demonstrating state-of-the-art performance and effective integration of real and simulated data for scalable, controllable video generation.
\end{itemize}

\section{Related Works}
\subsection{Controllable Video Generation}

Controllable video generation leverages diffusion models pretrained on large-scale video datasets~\cite{ho2020denoising, liu2022flow, ho2022video, singer2022make,blattmann2023stable} to synthesize videos under user-specified constraints. Spatial control methods, such as ControlNeXt~\cite{peng2024controlnext} and MimicMotion~\cite{zhang2024mimicmotion}, use masks and keypoints to guide object appearance and pose, while Champ~\cite{zhu2024champ} incorporates optical flow for motion control. Temporal control is addressed by Tora~\cite{zhang2025tora}, CameraCtrl~\cite{he2024cameractrl}, and MOFA-Video~\cite{niu2024mofa}, which introduce trajectory or camera motion cues for dynamic regulation. More recent efforts~\cite{li2024dispose, wang2024humanvid, luo2025dreamactor, wang2025multi, chen2025dancetogether, li2025building} attempt to combine spatial and temporal signals for finer-grained control, extending to multi-person interactive scenarios with identity preservation. Meanwhile, diffusion-based approaches have also been applied to articulated character animation~\cite{sun2025drive, song2025magicarticulate} and temporally consistent human-centric dense prediction~\cite{khirodkar2024sapiens,miao2026framessequencestemporallyconsistent, su2026interaction}, further broadening the scope of controllable generation. Despite the effectiveness of existing methods, they operate primarily on 2D representations, limiting their ability to capture complex 3D geometry and hindering realistic and scalable video synthesis. In this work, we introduce explicit 3D conditioning into video diffusion models while focusing on the specific hand-object interaction video generation task.

\subsection{Hand-Object Interaction Video Generation}

Hand-object interaction video generation encompasses both 3D reconstruction and 2D synthesis methods. 3D approaches, such as ARCTIC~\cite{fan2023arctic}, HOLD~\cite{fan2024hold}, ObMan~\cite{hasson2019learning}, and HOIDiffusion~\cite{zhang2024hoidiffusion}, focus on reconstructing or generating 3D hand-object poses, while Text2HOI~\cite{cha2024text2hoi} enables text-driven 3D motion synthesis. However, these methods typically operate on isolated objects without incorporating full scene context, which limits their applicability to real-world scenarios. In the 2D domain, CosHand~\cite{sudhakar2024controllingworldsleighthand} synthesizes static HOI images, and InterDyn~\cite{akkerman2025interdyn} as well as TASTE-Rob~\cite{zhao2025taste} extend generation to videos. Despite recent progress, 2D HOI generation still suffers from limited visual quality and physical inconsistency, which reduces its usefulness for downstream tasks such as robotic policy learning. We therefore study 3D-conditioned HOI generation to improve both visual fidelity and physical plausibility.

\subsection{3D Representation and Rendering}
3D rendering methods provide the foundation for synthesizing visual content from geometric data. Traditional graphics pipelines render meshes or point clouds via rasterization~\cite{yifan2019differentiable}, offering precise geometric control but requiring extensive manual asset preparation. Neural rendering techniques, such as NeRF~\cite{mildenhall2021nerf} and 3D Gaussian Splatting~\cite{kerbl20233d}, learn implicit or explicit 3D representations from multi-view images, enabling photorealistic novel view synthesis at the cost of scene-specific optimization. 
Recent works have also explored efficient 3D asset creation from diverse input modalities, including single-image reconstruction~\cite{li2024pshuman,chen2026ultraman,long2024wonder3d,ye2025hi3dgen,lei2025hunyuan3d}, interleaved multimodal 3D generation~\cite{chen2025idea23d,wang2024llama,weng2026garmentgpt}. In controllable image and video generation, most approaches convert 3D information into 2D rendered maps, such as depth~\cite{jang2025frame, pang2025dreamdance, peng2024controlnext, lapid2023gd}, surface normals~\cite{wang2025multi}, or pose visualizations~\cite{wang2024vividpose, li2024dispose, zhang2024mimicmotion}, to serve as conditioning signals for 2D diffusion models. While this rendering-then-generation paradigm is effective, it inevitably incurs information loss and struggles to capture complex 3D spatial relationships, especially under occlusion. In contrast, we directly incorporate 3D point clouds as conditioning signals, allowing the diffusion model to operate as a neural renderer and maintain full 3D structural information throughout generation.

\section{Methods}

\begin{figure*}
    \centering
    \includegraphics[width=\linewidth]{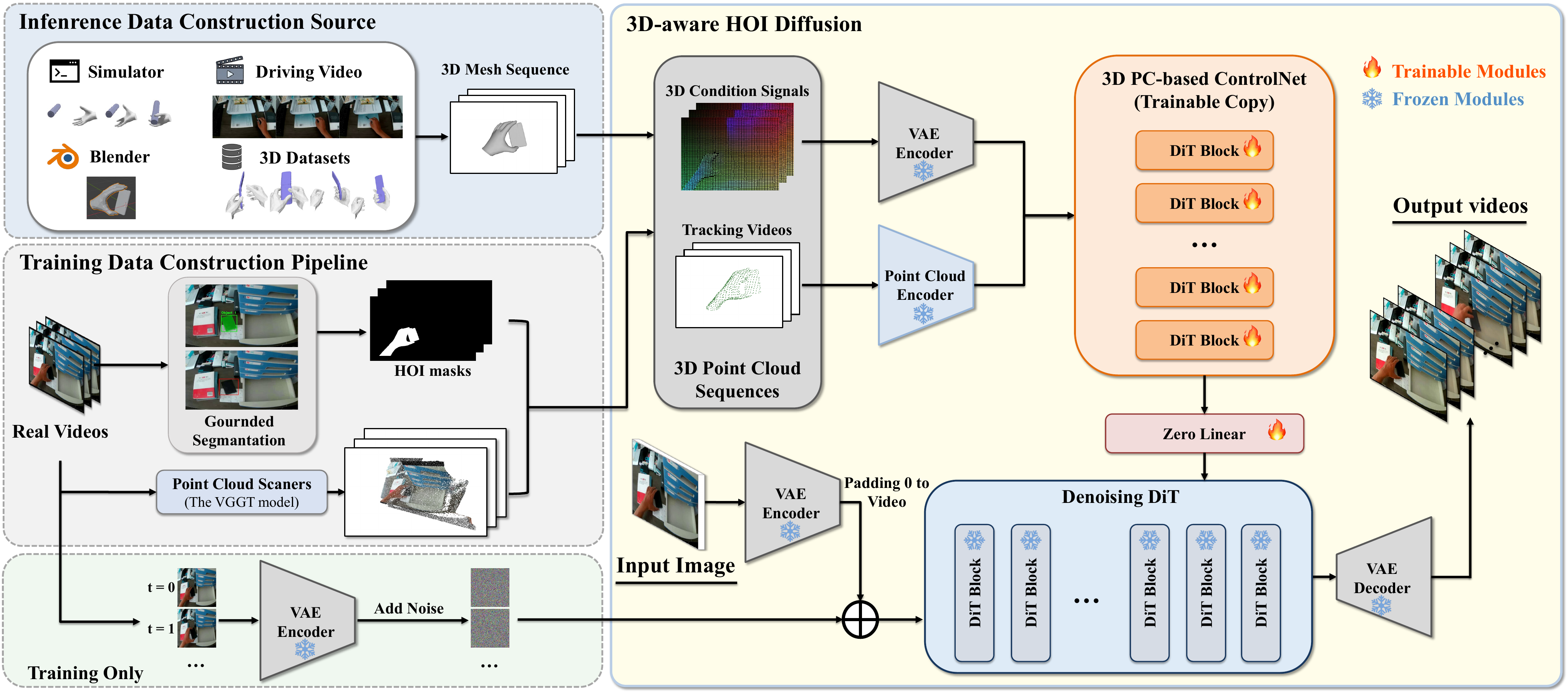}
    \caption{\textbf{Architecture of HVG-3D}. The left panel illustrates the hybrid training and inference pipeline, where egocentric driving videos, simulator outputs, and 3D HOI datasets are processed by grounded segmentation, key bounding-box extraction and a point-cloud scanner to construct paired input images, 3D tracking videos, and 3D point cloud sequences. The right panel depicts the 3D-aware HOI video generation diffusion architecture, in which the 3D point cloud and tracking signals are encoded by a trainable 3D ControlNet and injected into a frozen image-to-video diffusion backbone via zero-initialized layers, enabling the synthesis of temporally coherent videos that respect the underlying 3D hand–object interaction geometry.}
    \label{fig:pipeline}
\end{figure*}

\subsection{Overview}


We consider the task of 3D-conditioned hand-object interaction image-to-video (I2V) generation. Given a single input image $I_0 \in \mathbb{R}^{H \times W \times 3}$, a sequence of 3D point clouds $P = \{P_t\}_{t=1}^T$, $P_t \in \mathbb{R}^{N \times 3}$ representing hand-object geometry over $T$ frames, and an optional 3D tracking sequence $\mathcal{T} = \{T_t\}_{t=1}^T$, the goal is to generate a video $V_{out} = \{I_t\}_{t=1}^T$, $I_t \in \mathbb{R}^{H \times W \times 3}$ that is visually realistic, temporally coherent, and faithfully respects the 3D spatial constraints.

To address this problem, as shown in Fig. \ref{fig:pipeline},  our proposed HVG-3D framework consists of two main components:
(i) a 3D-aware diffusion-based architecture that encodes geometric and motion cues from the 3D point cloud and tracking sequence, enabling explicit 3D reasoning; and
(ii) a hybrid pipeline for constructing input and condition signals, which flexibly integrates both real and simulated data to provide precise spatial and temporal control for both training and inference.

In the following sections, we first detail the 3D-aware diffusion-based architecture (Section~\ref{3.2}), which builds upon a pretrained image-to-video diffusion backbone and incorporates 3D control signals. We then introduce the hybrid pipeline for constructing and aligning input and condition signals (Section~\ref{3.3}), including the acquisition and processing of 3D cues in both real and synthetic settings.

\subsection{3D-aware HOI Diffusion Architecture}
\label{3.2}

While recent diffusion-based image-to-video models achieve impressive visual quality, their reliance on 2D conditions limits spatial consistency and geometric fidelity, especially for hand–object interactions. To address this, we propose a 3D-aware diffusion framework that explicitly incorporates 3D structural and motion cues into the generation process. Our architecture consists of a strong image-to-video diffusion backbone and a dedicated 3D point cloud–guided ControlNet, as described below.

\noindent \textbf{Base Image-to-Video Model.}
Our framework adopts CogVideoX-5B-I2V~\cite{yang2024cogvideox} as the image-to-video generation backbone. CogVideoX-5B-I2V is a Transformer-based video diffusion model with 3D full attention, enabling high-fidelity and temporally consistent synthesis. As shown in Fig.\ref{fig:pipeline}, the model takes an input image $I_0 \in \mathbb{R}^{H \times W \times 3}$ and a ground-truth video $V_{gt} \in \mathbb{R}^{T \times H \times W \times 3}$, which are encoded into latent representations $Z_{I_0}$ and $Z_{gt} \in \mathbb{R}^{T \times \frac{H}{8} \times \frac{W}{8} \times 16}$ via a VAE encoder\cite{kingma2013auto}. The image latent $Z_{I_0}$ is temporally zero-padded to match the length $T$ and concatenated with a noised version of $Z_{gt}$. The resulting joint latent sequence is processed by a Diffusion Transformer, which performs iterative denoising to recover the clean video latent $Z_\varepsilon$. Finally, a 3D VAE decoder reconstructs the output video $V_{out}$ from $Z_\varepsilon$. Such a diffusion-based architecture provides robust temporal modeling and fine-grained control over video dynamics, facilitating the generation of realistic and structurally consistent hand–object interaction sequences. Its unified latent space enables effective encoding of both appearance and motion, supporting generalization to diverse HOI scenarios.

\noindent  \textbf{3D Point Cloud-based ControlNet.}
To provide explicit 3D structural and motion guidance, we introduce 3D point clouds as conditioning signals within our diffusion framework. Given a sequence of point clouds $P \in \mathbb{R}^{T \times N \times 3}$ extracted from the input video, we employ a point cloud encoder~\cite{zhang20233dshape2vecset} to obtain latent features $Z_{pc} \in \mathbb{R}^{T \times L \times 768}$, where $N$ denotes the number of points and $L$ the number of latent tokens. In parallel, 3D tracking information is encoded as $Z_{tracking} \in \mathbb{R}^{T \times \frac{H}{8} \times \frac{W}{8} \times 16}$.

To ensure compatibility among heterogeneous condition signals, we project $Z_{pc}$ via a learnable linear layer and resample to match the dimensionality of $Z_{tracking}$ and $Z_{gt}$. The aligned latents are concatenated and serve as input to the 3D Point Cloud ControlNet. Architecturally, the ControlNet is constructed by replicating all pretrained DiT blocks, which are then specialized to encode the 3D conditioning. At each layer, the output of the ControlNet is modulated by a zero-initialized convolution and injected into the corresponding DiT block of the main diffusion backbone. By injecting 3D structural and motion cues at every denoising step, our design significantly improves spatial consistency and physical plausibility in synthesized hand–object interactions, especially under challenging scenarios such as severe occlusions and complex articulations.

\subsection{Hybrid Pipeline for Input and Condition Signal Construction}
\label{3.3}

Our hybrid pipeline is designed to flexibly support both real and synthetic 3D conditioning signals throughout training and inference. It consists of three stages: training data construction, model training, and inference and  practical conditioning. Next, we detail each stages.

\noindent \textbf{Training Data Construction.} To address the lack of explicit mask and 3D point cloud annotations in Taste-Rob~\cite{zhao2025taste}, we devise a data pipeline for recovering these signals from monocular egocentric RGB videos of hand-object interaction. Object and hand bounding boxes are extracted by combining inter-frame difference maps (for static backgrounds) and YOLOv8-X~\cite{yolov8_ultralytics} detection (for dynamic hand regions).
Instance masks for both hand and object are generated using SAMURAI~\cite{yang2024samurai}, with tracking initialized via bounding boxes and bidirectional refinement to ensure temporal consistency. The masks are fused to obtain a per-frame hand-object segmentation. 3D point cloud reconstruction is performed with VGGT~\cite{wang2025vggt}: given the video frames and their corresponding masks, VGGT produces a per-frame point cloud $P \in \mathbb{R}^{T \times N \times 3}$ representing the 3D geometry of the hand and object.

\noindent \textbf{Model Training.}
With the training data constructed as described above, we proceed to optimize our 3D-aware diffusion model for hand-object interaction video generation. While explicit 3D conditioning provides strong geometric control, it may not fully suppress background distractions, especially in cluttered scenes. To address this, we augment the standard diffusion loss with a mask-based reconstruction term inspired by StableAnimator~\cite{tu2025stableanimator}, which focuses learning on the regions of interest.

The final training objective is defined as:
\begin{equation}
L = \sum_{i=1}^n \mathbb{E}\varepsilon \left( \left| \left(Z_{gt} - Z_\varepsilon \right) \odot (1 + M^i) \right|^2 \right)
\end{equation}
where $Z_{gt}$ and $Z_\varepsilon$ denote the ground-truth and predicted video latents respectively, and $M^i \in \{{0,1\}}^{1 \times H \times W}$ is the hand–object mask for frame $i$. This loss formulation ensures that errors in the hand-object regions are emphasized, promoting accurate reconstruction of interaction-critical areas while mitigating the influence of background noise.

For training, each video is center-cropped and resized to $720 \times 480$, with a fixed length of 49 frames. Each training sample comprises: the input image $I_0$, ground-truth video $V_{gt}$, the hand–object mask sequence, 3D point cloud sequence $P$, and 3D tracking sequence (estimated via SpatialTracker~\cite{xiao2024spatialtracker}). We fine-tune only the copied condition DiT blocks, keeping all parameters of the original denoising DiT backbone frozen to preserve pre-learned video generation capabilities. Training is performed using the AdamW optimizer with a learning rate of $1 \times 10^{-4}$ for 20 epochs, employing gradient accumulation to achieve an effective batch size of 4. All experiments are conducted on 8 H20 GPUs.

\noindent \textbf{Inference and Practical Conditioning.}
At inference time, the model takes a single real input image $I_0$ together with a 3D conditioning signal. The 3D condition can be: (1)extracted from real video using the same pipeline as in training (detection, segmentation, and VGGT point cloud reconstruction); or (2) synthesized in simulation, e.g., by generating 3D hand-object sequences in Blender or other simulators, or by sampling from 3D HOI datasets such as ARCTIC~\cite{fan2023arctic} or HOT3D~\cite{banerjee2024introducing}.
All 3D mesh sequences are processed to produce compatible point clouds and, if needed, tracking sequences, ensuring seamless integration with the model’s conditioning interface.

\section{Experiment}

\newcommand{\sotacell}[1]{\cellcolor{cyan!5}{\textbf{#1}}}
\begin{table*}[t]
\centering
\caption{\textbf{Quantitative comparison between HVG-3D and baselines on Full Frame evaluation metrics}. Most video generation metrics demonstrate that HVG-3D achieves superior performance.}
\renewcommand{\arraystretch}{1.15}

\resizebox{0.8\textwidth}{!}{%
\begin{tabular}{@{}l*{9}{c}@{}}
\toprule
Method &
L1$\downarrow$ &
PSNR$\uparrow$ &
SSIM$\uparrow$ &
LPIPS$\downarrow$ &
CLIP$\uparrow$ &
ST-SSIM$\uparrow$ &
GMSD-T$\downarrow$ &
FID$\downarrow$ &
C-FID$\downarrow$ \\
\midrule
Kling~\cite{kuaishou2024kling}      & 17.42 & 19.22 & 0.66 & 0.316 & 0.95 & 0.85 & 0.44 &  98.9 & 18.5 \\
Wan2.2~\cite{wan2025wan}            & 14.26 & 20.76 & 0.77 & 0.25  & 0.95 & 0.87 & 0.45 & 122.6 & 18.1 \\
CogVideoX~\cite{yang2024cogvideox}  & 22.85 & 17.10 & 0.67 & 0.334 & 0.93 & 0.77 & 0.45 & 174.2 & 27.6 \\
InterDyn~\cite{akkerman2025interdyn} &  8.81 & 24.13 & 0.82 & 0.205 & 0.95 & 0.95 & 0.45 &  73.3 & 17.8 \\
DAS~\cite{gu2025diffusion}          &
  \sotacell{7.77}  & \sotacell{24.83} & \sotacell{0.84} &
  \sotacell{0.191} & \sotacell{0.96} & 0.96 & 0.44 &
  75.5 & \sotacell{14.6} \\
\midrule
Our                                 &
  9.50 & 24.15 & 0.81 &
  \sotacell{0.193} &
  \sotacell{0.96} & \sotacell{0.97} & \sotacell{0.40} &
  \sotacell{58.2} & \sotacell{14.6} \\
\bottomrule
\end{tabular}%
}
\label{tab:full_frame}
\end{table*}

\begin{table*}[t]
\centering      
\caption{\textbf{Quantitative comparison between HVG-3D and baselines on Hand Object Masked Region evaluation metrics}. All video generation metrics consistently indicate that HVG-3D delivers superior performance.}
\renewcommand{\arraystretch}{1.15}

\resizebox{0.8\textwidth}{!}{%
\begin{tabular}{@{}l*{9}{c}@{}}
\toprule
Method &
L1$\downarrow$ &
PSNR$\uparrow$ &
SSIM$\uparrow$ &
LPIPS$\downarrow$ &
CLIP$\uparrow$ &
ST-SSIM$\uparrow$ &
GMSD-T$\downarrow$ &
FID$\downarrow$ &
C-FID$\downarrow$ \\
\midrule
Kling~\cite{kuaishou2024kling}    & 41.86 & 14.74 & 0.95 & 0.055 & 0.94 & 0.77 & 0.17 & 182.4 & 24.7 \\
Wan2.2~\cite{wan2025wan}          & 48.36 & 13.77 & 0.95 & 0.061 & 0.94 & 0.72 & 0.18 & 217.1 & 27.1 \\
CogVideoX~\cite{yang2024cogvideox}& 58.78 & 11.83 & 0.94 & 0.068 & 0.92 & 0.62 & 0.20 & 260.9 & 36.9 \\
InterDyn~\cite{akkerman2025interdyn}
                                  & 20.99 & 19.03 &
                                    0.96 & 0.034 & 0.96 &
                                    0.92 & 0.16 &
                                    104.5 & 14.5 \\
DAS~\cite{gu2025diffusion}        & 26.55 & 17.41 & 0.96 & 0.039 & 0.96 & 0.88 & 0.16 & 128.0 & 16.0 \\
\midrule
Our                               &
  \sotacell{20.90} & \sotacell{19.08} & \sotacell{0.97} &
  \sotacell{0.032} & \sotacell{0.97} & \sotacell{0.93} &
  \sotacell{0.15} & \sotacell{88.5} & \sotacell{13.1} \\
\bottomrule
\end{tabular}%
}
\label{tab:mask}
\end{table*}

\subsection{Experiment Setting}
\textbf{Datasets} We train our HVG-3D on Taste-Rob~\cite{zhao2025taste}. This datasset is an egocentric hand object dataset. This dataset comprises egocentric videos of hand–object interactions collected across multiple scenes. For training, we selected the Single Hand subset and used samples with scene labels {office, dining, bedroom, kitchen, dressing table}. We crop all videos to a resolution of 720×480 and segment the original videos into clips of 49 frames each. In the evaluation stage, we first sampled 2\% of the videos from each scene category as candidate test samples. For the Taste-Rob evaluation, we then randomly selected 100 videos from these candidates to construct our final test set. All evaluation metrics are reported based on this test set.

\textbf{Metrics.} We evaluate the performance of our model from two complementary aspects: \textit{image quality} and \textit{spatio-temporal similarity}. \textbf{Image quality.} To assess the fidelity of the generated frames, we adopt several commonly used image-based metrics, including L1, Peak Signal-to-Noise Ratio (PSNR), the Structural Similarity Index Measure (SSIM)~\cite{1284395}, Learned Perceptual Image Patch Similarity (LPIPS), CLIP Score~\cite{hessel2021clipscore}, Fréchet Inception Distance (FID)~\cite{NIPS2017_8a1d6947}, and CLIP-FID. These metrics comprehensively measure the perceptual and structural consistency between generated and ground-truth frames. \textbf{Spatio-temporal similarity.} To further evaluate the overall video quality across both spatial and temporal dimensions, we measure the perceptual similarity between the generated and real video distributions using the Fréchet Video Distance (FVD)~\cite{unterthiner2019fvd}, Spatio-Temporal SSIM(ST-SSIM)~\cite{moorthy2010efficient} and Gradient Magnitude Similarity Deviation – Temporal(GMSD-T)~\cite{yan2015video}. These metrics quantify the coherence and realism of motion dynamics in the generated videos.

\subsection{Baseline Comparisons}
To ensure a comprehensive and fair comparison, we select three state-of-the-art video generation models, namely CogVideoX~\cite{yang2024cogvideox}, Wan 2.2~\cite{wan2025wan}, Kling~\cite{kuaishou2024kling} and DaS~\cite{gu2025diffusion}. In addition, we compare our method with a specialized approache for hand–object interaction video generation, namly, InterDyn~\cite{akkerman2025interdyn}.

\textbf{Quantitative comparison.} To compare HVG-3D with existing methods, we divide each video in the test set into 49-frame clips and randomly select one clip containing hand–object interaction from each video for evaluation. For each baseline method, we extract the required conditional inputs from its corresponding video.

\begin{figure}[t]
    \centering
    \includegraphics[width=0.95\columnwidth]{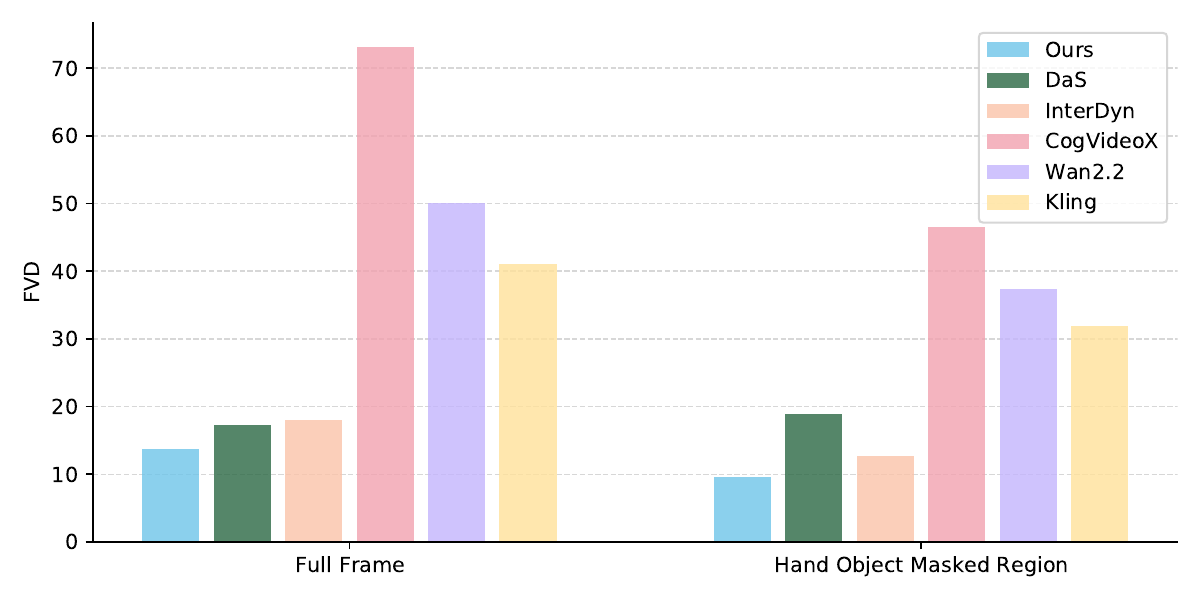}
    \caption{\textbf{Qualitative comparison between HVG-3D and baselines on FVD}. Our method achieves the best FVD scores in both the full-frame setting and the hand–object masked region.}
    \label{fig:fvd_bar}
\end{figure}

As shown in Tab.~\ref{tab:full_frame} and Tab~\ref{tab:mask}, Tab.~\ref{tab:full_frame} reports the full-frame performance metrics, whereas Tab~\ref{tab:mask} presents the mask-aware metrics within the hand–object region. Under full-frame evaluation, benefiting from the precise control provided by the 3D condition, our method achieves the lowest FVD (13.8) and FID (58.2), while also obtaining the highest CLIP Score (0.96) and GMSD-T (0.40). Although some full-frame metrics are slightly inferior to those of DaS, our method attains the best performance on all metrics within the hand–object mask region, which corresponds to the primary interaction area. In particular, FVD is reduced from 13.8 to 9.6 as shown in the Fig.~\ref{fig:fvd_bar}, and C-FID decreases from 14.6 to 13.1, even as other methods exhibit a consistent degradation across all metrics in this region. Notably, these improvements are achieved without sacrificing low-level reconstruction fidelity. L1 and LPIPS are simultaneously reduced, whereas PSNR and SSIM are improved.

\begin{figure*}[p]
    \centering
    \includegraphics[height=0.96\textheight,keepaspectratio]{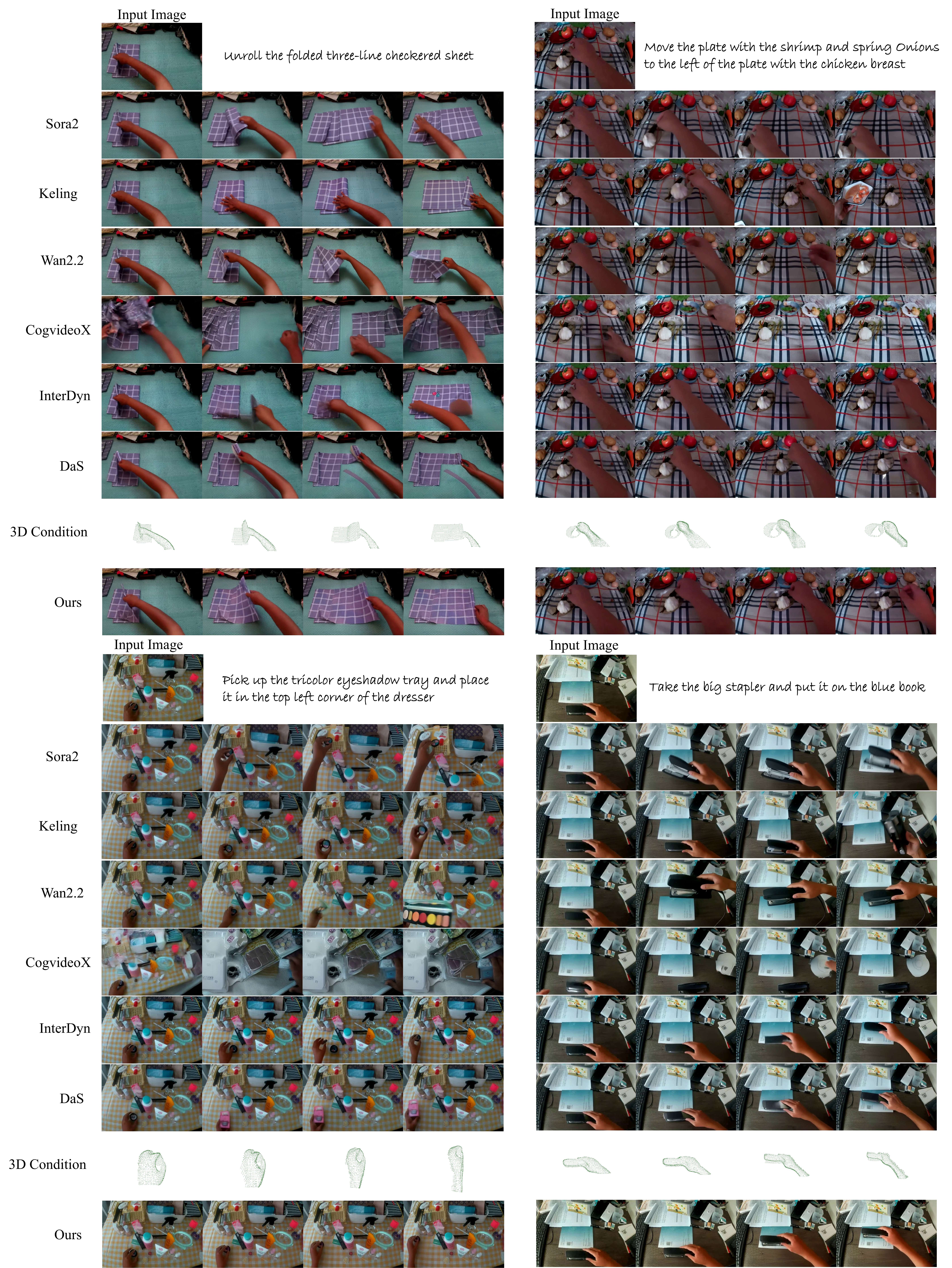}
    \caption{\textbf{Qualitative comparison of video generation performance}. HVG-3D is capable of generating videos with highly accurate motions and superior visual quality, while further ensuring that both the hand and the object remain free from geometric deformation. A level of performance that current state-of-the-art general-purpose video generation models are unable to achieve.}
    \label{fig:qualitative}
\end{figure*}

\textbf{Qualitative comparison.} 
The qualitative comparison is presented in Fig.~\ref{fig:qualitative}. In the first case, we illustrate the process of unfolding a folded three-line checkered sheet. In the second case, we show a plate containing shrimp and scallions being moved to the left side of a plate with chicken breast. In the third case, we demonstrate placing a three-tone eyeshadow palette at the upper-left corner of a dressing table. In the fourth case, we depict placing a stapler on top of a blue book. For each example, we provide the input image, the text prompt, and the method-specific conditions (the tracking video for DaS and the mask for InterDyn). Moreover, since the output resolution of Sora2 is not aligned with that of the other baselines, a fair quantitative comparison is not feasible. We therefore only present qualitative results in this section.

As shown in Fig.~\ref{fig:qualitative}, only our method, HVG-3D, successfully executes the specified manipulation while preserving the shape of both the hand and the object. DaS performs reasonably well for in-plane parallel translations of the object, but once the task involves folding or motion perpendicular to the tabletop, it tends to introduce noticeable object deformation. InterDyn exhibits similar issues and is even less stable than DaS. In contrast, the recent powerful video generation models CogVideoX, Wan2.2, Kling, and Sora2 all struggle to reliably accomplish the required manipulation.

Beyond the aforementioned qualitative results, we further demonstrate the flexibility of HVG-3D in handling conditioning sources at inference time. As shown in Fig.~\ref{fig:pipeline}, our framework not only accepts 3D point clouds scanned from videos, but also ingests 3D conditions derived from diverse pipelines, including physics-based simulators, real driving videos with 3D reconstruction, and pre-captured hand–object interaction datasets. In practical deployments, HVG-3D can be driven by a broad range of 3D inputs, including 3D mesh sequences edited in Blender to create novel hand–object motions, 3D trajectories or point clouds estimated from driving videos through standard reconstruction pipelines, ready-made 3D conditions from existing hand–object interaction datasets, and synthetic 3D hand–object interaction sequences rendered directly by simulators. As shown in the last two rows of Fig.~\ref{fig:teaser}, we further demonstrate that editing 3D mesh sequences in Blender enables the generation of new hand–object interaction videos. This unified interface for heterogeneous 3D inputs underscores the generality of our framework and supports seamless deployment across diverse 3D acquisition pipelines and downstream application scenarios.

\subsection{Ablation Study}
In this section, we present ablation studies to validate the effectiveness of our 3D point-cloud conditioning. Our ablation studies further demonstrate that combining the 3D tracking video with the proposed mask diffusion loss not only improves manipulation accuracy in human–object interaction scenarios and better preserves the shapes of the hand and the object, but also accelerates training and enables faster model convergence.

\textbf{Ablations on 3D point cloud condition} 3D point-cloud conditioning equips the model with more accurate 3D perception during training, thereby strengthening depth reasoning. To assess its contribution, we remove this condition and train the model for the same number of epochs as the full system, then compare the results. As shown in Tab.~\ref{tab:ablation}, omitting the 3D point cloud condition degrades the quality of hand–object interactions in the generated videos. The deterioration of these metrics further reflects that the absence of the 3D point cloud condition leads to noticeable shape distortions of the hand and object during interaction. Moreover, similar to the phenomenon observed in DaS, when the object needs to be folded or undergoes vertical motion perpendicular to the tabletop, meaning that the model must demonstrate a certain degree of depth awareness, the quality of the generated videos decreases substantially. These factors collectively contribute to the lower evaluation scores.


\begin{table}[t]
\centering
\caption{Ablation Studies on 3D point cloud, 3D tracking video and mask diffusion loss. The experimental results demonstrate that these techniques enhance the quality of hand–object interaction video generation, improve the accuracy of the synthesized interaction process, and accelerate convergence during training.}
\label{tab:ablation}
\renewcommand{\arraystretch}{1.15}
\resizebox{0.8\columnwidth}{!}{%
\begin{tabular}{@{}lcccc@{}}
\toprule
Method & PSNR & SSIM & LPIPS & ST-SSIM \\ 
\midrule
w/o 3D pc       & 18.44 & 0.37 & 0.5957 & 0.91 \\
w/o 3D tracking & 22.76 & 0.75 & 0.2054 & 0.95 \\
w/o mask loss   & 22.09 & 0.80 & 0.199  & 0.96 \\
full model      & \textbf{24.15} & \textbf{0.81} & \textbf{0.193} & \textbf{0.97} \\ 
\bottomrule
\end{tabular}%
}
\end{table}

\textbf{Ablations on 3D tracking video} 3D tracking videos provide the model with more accurate viewpoint awareness and, when combined with 3D point-cloud conditioning, enhance control over hand–object interactions. To evaluate their effectiveness, we ablate the 3D tracking video condition, train for the same number of epochs as the full model, and then compare results. As shown in Tab.~\ref{tab:ablation}, the metrics indicate that removing the 3D tracking video leads to a degradation in the quality of hand–object interactions in the generated sequences. In the absence of 3D tracking video, the object shape may remain plausible, but the spatial alignment of the interaction (e.g., contact locations and motion trajectories) becomes less accurate. Moreover, since the model is trained with the 3D point cloud condition but without the 3D tracking video, the resulting performance drop suggests that the 3D tracking video encodes complementary camera-view information that helps the model learn more effective point cloud representations.


\textbf{Ablations on mask diffusion loss} Mask diffusion loss encourages the model to focus on hand–object interaction regions during training, thereby improving convergence. To assess its effectiveness, we remove the mask component from the diffusion loss and train for the same number of epochs as the full model, then compare the results. As shown in Tab.~\ref{tab:ablation}, removing the mask diffusion loss leads to a degradation in overall video quality. This occurs because, once the mask is excluded from the diffusion loss, the model tends to focus on the entire scene during training rather than emphasizing the hand–object interaction region. Under the same training configuration as the full model, the metrics obtained at the same number of epochs are consistently worse, indicating that explicitly guiding the model to focus on hand–object interactions enables faster training and more rapid convergence. At the same time, the model becomes more robust to distractions from other objects in complex scenes, resulting in more accurate generation of hand–object interactions.

\section{Conclusion}
We presented HVG-3D, a unified framework for 3D-conditioned hand–object interaction video generation. By incorporating a 3D ControlNet that encodes point cloud and tracking cues into a video diffusion backbone, together with a hybrid pipeline bridging real and simulated domains, HVG-3D achieves state-of-the-art spatial fidelity, temporal coherence, and controllability on the TASTE-Rob benchmark. Ablation studies confirm the complementary benefits of each component. Future work will extend to more diverse interaction scenarios, longer sequences, and closed-loop integration with robotic manipulation policies.


\section{Acknowledgements}
This work was supported by the New Generation Artificial Intelligence-National Science and Technology Major Project (2025ZD0122603). It was also supported by the Postdoctoral Fellowship Program and China Postdoctoral Science Foundation under Grant No. 2024M764093 and Grant No. BX20250485, the Beijing Natural Science Foundation under Grant No. 4254100, and by Beijing Advanced Innovation Center for Future Blockchain and Privacy Computing. It was also supported by the Young Elite Scientists Sponsorship Program of the Beijing High Innovation Plan (NO. 20250860).

The research work described in this paper was conducted in the JC STEM Lab of Machine Learning and Computer Vision funded by The Hong Kong Jockey Club Charities Trust. This research received partially support from the Global STEM Professorship Scheme from the Hong Kong Special Administrative Region. 


{
    \small
    \bibliographystyle{ieeenat_fullname}
    \bibliography{main}
}

\end{document}